\documentclass[letterpaper, 10 pt, conference]{ieeeconf}
\IEEEoverridecommandlockouts  
\usepackage{bm}
\usepackage{amssymb}
\usepackage{amsthm}
\usepackage{mathtools}
\usepackage{nicefrac}  

\usepackage[noend]{algpseudocode}
\usepackage{algorithm}
\usepackage{setspace}

\usepackage{booktabs}
\usepackage{multicol}
\usepackage{multirow}
\usepackage{tabularx}
\usepackage{etoolbox}
\newrobustcmd{\B}{\bfseries}  
\usepackage{siunitx}  
\usepackage{diagbox}
\usepackage{pifont}  
\newcommand{\cmark}{\ding{51}}%
\usepackage{todo}  
\usepackage{xcolor}
\definecolor{red}{RGB}{200,50,50}
\definecolor{green}{RGB}{50,200,50}
\definecolor{blue}{RGB}{50,50,200}
\definecolor{orange}{RGB}{243,147,82}
\definecolor{purple}{RGB}{150,100,250}
\definecolor{yellow}{RGB}{231,198,100}
\definecolor{gray}{RGB}{98,95,75}
\definecolor{white}{RGB}{255,255,255}
\usepackage{lipsum}
\usepackage{times}

\makeatletter
\newcommand*{\T}{{\mathpalette\@transpose{}} }
\newcommand*{\@transpose}[2]{\raisebox{\depth}{$\m@th#1\intercal$}}
\makeatother
\DeclarePairedDelimiterX{\norm}[1]{\lVert}{\rVert_{2}}{#1}

\makeatletter
\newcommand\thefontsize{The current font size is: \f@size pt}
\makeatother

\makeatletter
\let\NAT@parse\undefined
\makeatother
\usepackage[numbers,sort&compress,square,comma]{natbib}  
\usepackage{hyperref}
\hypersetup{
  colorlinks=true,
  urlcolor=blue,
}
\usepackage{cleveref}  
\theoremstyle{definition}
\newtheorem{problem}{Problem}
\theoremstyle{definition}

\theoremstyle{definition}
\newtheorem{definition}{Definition}
\theoremstyle{definition}

\theoremstyle{definition}

\crefname{lemma}{Lemma}{Lemmas}
\crefname{proposition}{Proposition}{Propositions}
\crefname{definition}{Definition}{Definitions}
\crefname{theorem}{Theorem}{Theorems}
\crefname{conjecture}{Conjecture}{Conjectures}
\crefname{corollary}{Corollary}{Corollaries}
\crefname{section}{Section}{Sections}
\crefname{appendix}{Appendix}{Appendices}
\crefname{figure}{Fig.}{Figs.}
\crefname{equation}{Eq.}{Eqs.}
\crefname{table}{Table}{Tables}
\crefname{algocf}{Algorithm}{Algorithms}
\crefname{problem}{Problem}{Problems}
\usepackage{balance}  
\apptocmd{\sloppy}{\hbadness 10000\relax}{}{}

\usepackage{graphicx}
\usepackage[caption=false,font=footnotesize]{subfig}
\usepackage{wrapfig}  
\usepackage{tikz}
\usetikzlibrary{positioning,shapes,bayesnet,calc,backgrounds}

\makeatletter
\newif\if@showgrid@grid
\newif\if@showgrid@left
\newif\if@showgrid@right
\newif\if@showgrid@below
\newif\if@showgrid@above
\tikzset{%
  every show grid/.style={},
  show grid/.style={execute at end picture={\@showgrid{grid=true,#1}}},%
  show grid/.default={true},
  show grid/.cd,
  labels/.style={font={\sffamily\small},help lines},
  xlabels/.style={},
  ylabels/.style={},
  keep bb/.code={\useasboundingbox (current bounding box.south west) rectangle (current bounding box.north west);},
  true/.style={left,below},
  false/.style={left=false,right=false,above=false,below=false,grid=false},
  none/.style={left=false,right=false,above=false,below=false},
  all/.style={left=true,right=true,above=true,below=true},
  grid/.is if=@showgrid@grid,
  left/.is if=@showgrid@left,
  right/.is if=@showgrid@right,
  below/.is if=@showgrid@below,
  above/.is if=@showgrid@above,
  false,
}
\def\@showgrid#1{%
  \begin{scope}[every show grid,show grid/.cd,#1]
    \if@showgrid@grid
      \begin{pgfonlayer}{background}
        \draw [help lines]
        (current bounding box.south west) grid
        (current bounding box.north east);
        \pgfpointxy{1}{1}%
        \edef\xs{\the\pgf@x}%
        \edef\ys{\the\pgf@y}%
        \pgfpointanchor{current bounding box}{south west}
        \edef\xa{\the\pgf@x}%
        \edef\ya{\the\pgf@y}%
        \pgfpointanchor{current bounding box}{north east}
        \edef\xb{\the\pgf@x}%
        \edef\yb{\the\pgf@y}%
        \pgfmathtruncatemacro\xbeg{ceil(\xa/\xs)}
        \pgfmathtruncatemacro\xend{floor(\xb/\xs)}
        \if@showgrid@below
          \foreach \X in {\xbeg,...,\xend} {
            \node [below,show grid/labels,show grid/xlabels] at (\X,\ya) {\X};
          }
        \fi
        \if@showgrid@above
          \foreach \X in {\xbeg,...,\xend} {
            \node [above,show grid/labels,show grid/xlabels] at (\X,\yb) {\X};
          }
        \fi
        \pgfmathtruncatemacro\ybeg{ceil(\ya/\ys)}
        \pgfmathtruncatemacro\yend{floor(\yb/\ys)}
        \if@showgrid@left
          \foreach \Y in {\ybeg,...,\yend} {
            \node [left,show grid/labels,show grid/ylabels] at (\xa,\Y) {\Y};
          }
        \fi
        \if@showgrid@right
          \foreach \Y in {\ybeg,...,\yend} {
            \node [right,show grid/labels,show grid/ylabels] at (\xb,\Y) {\Y};
          }
        \fi
      \end{pgfonlayer}
      \fi
    \end{scope}
  }
  \makeatother
  \tikzset{every show grid/.style={show grid/keep bb}%
  }%

\title{\LARGE\bf Model-Agnostic Multi-Agent Perception Framework}

\author{Runsheng Xu$^{1*}$, Weizhe Chen$^{2*}$, Hao Xiang$^{1*}$, Xin Xia$^{1}$, Lantao Liu$^{2}$, Jiaqi Ma$^{1}$
\thanks{*The first three authors contribute equally.}
\thanks{$^{1}$Runsheng Xu, Hao Xiang, Xin Xia, and Jiaqi Ma are with the University of California, Los Angeles, CA, USA \{rxx3386,haxiang,x35xia\}@g.ucla.edu, jiaqima@ucla.edu}%
\thanks{$^{2}$Weizhe Chen and Lantao Liu are with Indiana University, Bloomington, IN, USA \{chenweiz, lantao\}@iu.edu}%
}

\begin{document}
\maketitle
\thispagestyle{empty}
\pagestyle{empty}
\begin{abstract}
Existing multi-agent perception systems assume that every agent utilizes the same model with identical parameters and architecture. The performance can be degraded with different perception models due to the mismatch in their confidence scores. In this work, we propose a model-agnostic multi-agent perception framework to reduce the negative effect caused by the model discrepancies without sharing the model information. Specifically, we propose a confidence calibrator that can eliminate the prediction confidence score bias. Each agent performs such calibration independently on a standard public database to protect intellectual property. We also propose a corresponding bounding box aggregation algorithm that considers the confidence scores and the spatial agreement of neighboring boxes. Our experiments shed light on the necessity of model calibration across different agents, and the results show that the proposed framework improves the baseline 3D object detection performance of heterogeneous agents. The code can be found at \href{https://github.com/DerrickXuNu/model_anostic}{this url}.
\end{abstract}
\section{Introduction}%
\label{sec:introduction}
Recent advancements in deep learning have improved the performance of modern perception systems on many tasks, such as object detection~\cite{zhou2018voxelnet, lu2021raanet, fan2021deep}, semantic segmentation ~\cite{pan2020cross, xiong2019adaptive}, and visual navigation~\cite{du2020learning, pal2021learning,9998013}. Despite the remarkable progress, single-agent perception systems still have many limitations due to single-view constraints. For instance, autonomous vehicles~(AVs) usually suffer from occlusion~\cite{liu2020vision, hua2019hierarchical, chen2023dynamic}, and such situations are difficult to handle because of the lack of sensory observations of the occluded area. To address this issue, recent studies~\cite{wang2020v2vnet, li2021learning, xu2022opencood, xu2022v2xvit, xiang2022v2xp,xu2022bridging} have explored wireless communication technology to enable nearby agents to share the sensory information and collaboratively perceive the surrounding environment. 

Although existing fusion frameworks have obtained a significant 3D object detection performance boost, they assume that all the collaborating agents share an identical model with the same parameters. This assumption is hard to satisfy in practice, particularly in autonomous driving. Distributing the model parameters among AVs might raise privacy and confidentiality concerns, especially for vehicles from different automotive companies. Even for AVs from the same company, the detection models can have various versions, depending on the vehicle type and model updating frequency. Without adequately handling the inconsistency, the shared sensory information can have a large domain gap, and the advantage brought by multi-agent perception will be diminished rapidly. 

\begin{figure}[t]%
  \centering%
  \subfloat{%
    \resizebox{\linewidth}{!}{
      \scriptsize
      \begin{tikzpicture}%
        \begin{scope}
        \clip [rounded corners=.5cm] (-3.4,-1.8) rectangle coordinate (centerpoint) (3.4,1.8);
        \node at(0.0,0.0){\includegraphics[width=0.8\linewidth]{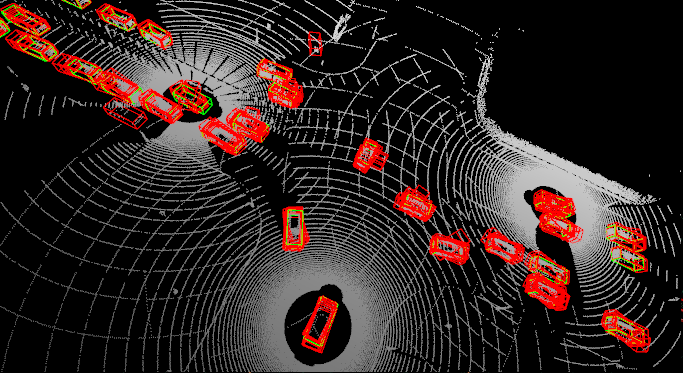}};%
        \end{scope}
        \begin{scope}
        \clip (-2,-3)  circle (1cm) ;
        \node[anchor=center] at (-2,-3) {\includegraphics[width=3cm]{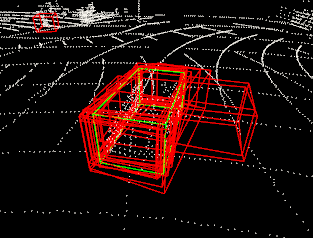}};
        \end{scope}
        \begin{scope}
        \clip (2,-3)  circle (1cm) ;
        \node[anchor=center] at (2,-3) {\includegraphics[width=2.7cm]{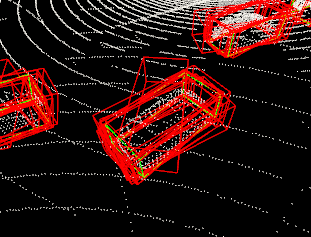}};
        \end{scope}
        \draw [-stealth,red](0.2,0.2) -- (-1.4,-2.2);
        \draw [-stealth,red](1.1,-0.8) -- (1.4,-2.2);
        \node at (0.5,-2.2) {\color{blue}$0.81$};
        \draw [-stealth,blue] (0.8,-2.2) -- (1.8,-2.5);
        \node at (0.5,-2.7) {\color{orange}$0.79$};
        \draw [-stealth,orange] (0.8,-2.7) -- (1.5,-3.0);
        \node at (0.5,-3.1) {\color{orange}$0.78$};
        \draw [-stealth,orange] (0.8,-3.1) -- (1.5,-3.1);
        \node at (0.5,-3.5) {\color{orange}$0.79$};
        \draw [-stealth,orange] (0.8,-3.5) -- (1.5,-3.2);
        \node at (0.5,-3.9) {\color{orange}...};
        \node at (-0.5,-2.2) {\color{blue}$0.81$};
        \draw [-stealth,blue] (-0.8,-2.2) -- (-1.1,-2.6);
        \node at (-0.5,-2.6) {\color{blue}$0.78$};
        \draw [-stealth,blue] (-0.8,-2.6) -- (-1.1,-3.0);
        \node at (-0.5,-3.1) {\color{orange}$0.71$};
        \draw [-stealth,orange] (-0.8,-3.1) -- (-1.8,-3.5);
        \node at (-0.5,-3.5) {\color{orange}$0.68$};
        \draw [-stealth,orange] (-0.8,-3.5) -- (-1.9,-3.6);
        \node at (-0.5,-3.9) {\color{orange}...};
        \node at (-2.0,-4.2) {(a)};
        \node at (2.0,-4.2) {(b)};
      \end{tikzpicture}%
    }%
  } \vspace{-10pt}
  \caption{\textbf{Ground truth ({\color{green}green}) and bounding box candidates ({\color{red}red}) produced by three connected autonomous vehicles}. (a) Some agents have confidence scores that are systematically larger than others, e.g., the {\color{blue}blue} scores versus the {\color{orange}orange} scores. However, they might be confidently wrong, which mislead the fusion process. (b) Candidates with slightly lower confidence scores ({\color{orange}orange}) but higher spatial agreement with neighboring boxes can be better than a singleton with a higher confidence score ({\color{blue}blue}). \vspace{-10pt}
  }\label{fig:pull_figure} 
\end{figure}
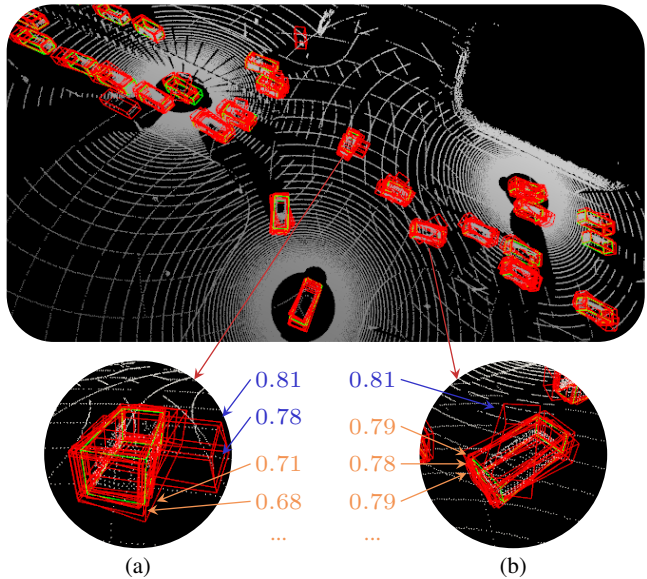


To this end, we propose a model-agnostic multi-agent perception framework to handle the model heterogeneity while maintaining confidentiality. The perception outputs (i.e., detected bounding boxes and confidence scores) are shared to bypass the dependency on the underlying model's detailed information. Due to the distinct models used by the agents, the confidence scores provided by different agents can be systematically misaligned. Some agents may be over-confident, whereas others tend to be under-confident. Directly fusing bounding box proposals from neighboring agents using, for example, Non-Maximum Suppression~(NMS)~\cite{neubeck2006efficient} can result in poor detection accuracy due to the presence of over-confident and low-quality candidates.

We propose a simple yet flexible confidence calibrator, called Doubly Bounded Scaling~(DBS), to mitigate the misalignment. We also propose a corresponding bounding box aggregation algorithm, named Promote-Suppress Aggregation~(PSA), that considers the confidence scores and the spatial agreement of neighboring boxes.
\cref{fig:pull_figure} illustrates the importance of these two components.
This framework does not reveal model design and parameters, ensuring confidentiality.
We evaluate our approach on an open-source large-scale multi-agent perception dataset -- OPV2V~\cite{xu2022opencood}. Experiments show that in the presence of model discrepancies among agents, our framework significantly improves multi-agent LiDAR-based 3D object detection performance, outperforming the baselines by at least 6\% in terms of Average Precision~(AP).

\section{Related Work}%
\label{sec:related_work}
\textbf{Multi-Agent Perception.}
Multi-agent systems have been extensively studied recently because of their potential to revolutionize robotics industry~\cite{9352029,9495943,9684293,9851671,9998013,khalil2022licanet,kitajima2022nationwide,shet2021cooperative,ferrara2022multi}. Multi-agent perception, as an important branch in multi-agent systems, investigates how to leverage visual cues from neighboring agents through the communication system to enhance the perception capability. There are three categories of existing work according to the information sharing schema: 1) early fusion, where raw point clouds are transmitted directly and projected into the same coordinate frame, 2) late fusion~\cite{rawashdeh2018collaborative}, where detected bounding boxes and confidence scores are shared, and 3) intermediate fusion~\cite{li2021learning, xu2022opencood, wang2020v2vnet, xu2022v2xvit, su2022uncertainty}, where compressed latent neural features extracted from point clouds are propagated. Though early fusion has no information loss, it usually requires large bandwidth. Intermediate fusion can achieve a good balance between accuracy and transmission data size, but it requires complete knowledge of each agent's model, which is non-trivial to satisfy in reality due to intellectual property concerns. On the contrary, late fusion only needs the outputs of the detector without demanding access to the underlying neural networks, which are typically confidential for automotive companies. Therefore, our approach adopts the late fusion strategy but further designs customized new components to address the model discrepancy issue in vanilla late fusion.


\textbf{Confidence Calibration.}
For a probabilistic classifier, the probability associated with the predicted class label should reflect its correctness likelihood. However, many modern neural networks do not have such property~\cite{guo2017calibration}. Confidence calibration aims to endow a classifier with such property. Calibration methods can be tightly coupled with the neural networks, such as Bayesian neural networks and regularization techniques~\cite{maddox2019simple,gal2016dropout,thulasidasan2019mixup}, or serve as a post-processing step. Post-processing methods include histogram binning methods~\citep{zadrozny2001obtaining}, scaling methods~\citep{platt1999probabilistic,zadrozny2002transforming}, and mixtures~\citep{kumar2019verified} that combine the first two branches. Due to the popularity of the Temperature Scaling method~\cite{guo2017calibration} which is a single-parameter version of Platt Scaling~\cite{platt1999probabilistic}, scaling methods are widely adopted for calibrating neural networks. Our proposed method follows the same fashion.

\textbf{Bounding Box Aggregation.}
Object detection models typically require bounding box aggregation to lump the proposals corresponding to the same object. Recent study~\cite{liu2022yolov5} demonstrates that bounding box aggregation can effectively improve small object detection accuracy. The \textit{de facto} standard post-processing method is Non-Maximum Suppression~(NMS)~\citep{neubeck2006efficient}, which sequentially selects the proposals with the highest confidence score and then suppresses other overlapped boxes. NMS does not fully exploit information in the proposals because it only uses the relative order of confidence, ignoring the absolute confidence scores and the spatial information hidden in the bounding box coordinates. Several works have been proposed to refine the box aggregation strategies. Soft-NMS~\citep{bodla2017soft} softly decays the confidence scores of the proposals proportional to the degree of overlap. In \cite{hosang2017learning} NMS can be learned by a neural network to achieve better occlusion handling and bounding box localization. Adaptive NMS~\cite{liu2019adaptive} applies a dynamic suppression threshold to an instance according to the target object density. \citet{rothe2014non} formulate NMS as a clustering problem and use Affinity Propagation Clustering to solve the problem. The idea of message passing between proposals is related to the aggregation algorithm introduced in \cref{sec:PSA}, but our update rules are simpler and more efficient.

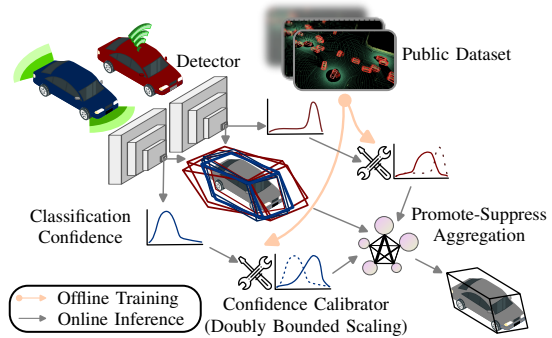
\begin{figure}[t]
  \centering
  \subfloat{%
      \begin{tikzpicture}\scriptsize%
        \node at(0.0,0.0){\includegraphics[width=0.8\linewidth]{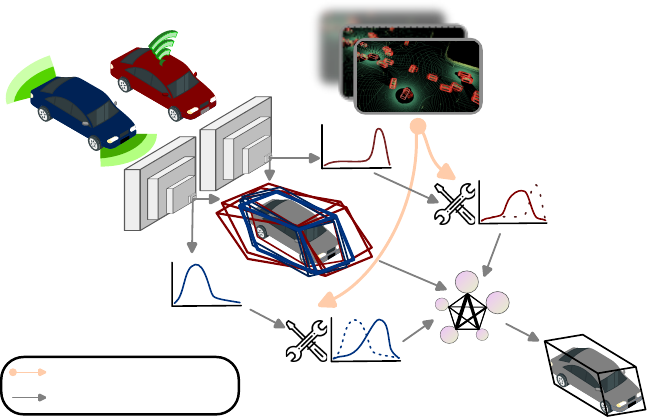}};%
        \node at (2.5,1.5) {Public Dataset};
        \node at (-0.8,1.4) {Detector};
        \node [align=center] at (-2.5,-0.8) {Classification\\Confidence};
        \node [align=center] at (0.5,-2) {Confidence Calibrator\\(Doubly Bounded Scaling)};
        \node [align=center] at (2.8,-0.8) {Promote-Suppress\\Aggregation};
        \node [align=left] at (-2,-1.78) {Offline Training};
        \node [align=left] at (-1.96,-2.03) {Online Inference};
      \end{tikzpicture}%
  }\vspace{-10pt}%
  \caption{\textbf{Overview of the proposed framework}. Each agent trains its confidence calibrator (i.e., Doubly Bounded Scaling) on the same public dataset offline (\textcolor{orange}{orange} arrows). Promote-Suppress Aggregation yields the final detection result, considering the spatial information and calibrated confidence of bounding boxes given by connected autonomous vehicles.\vspace{-10pt}}
  \label{fig:architecture}
\end{figure}

\section{Methodology}%
\label{sec:methodology}
In this paper, we consider the cooperative perception of a heterogeneous multi-agent system, where agents communicate to share sensing information from different perception models without revealing model information, i.e., model-agnostic collaboration. We focus on a 3D LiDAR detection task in autonomous driving, but the methodology can also be customized and used in other cooperative perception applications. Our goal is to develop a robust framework to handle the heterogeneity among agents while preserving confidentiality. The proposed model-agnostic collaborative perception framework is shown in ~\cref{fig:architecture}, which can be divided into two stages. In the offline stage, we train a model-specific calibrator. During the online phase, real-time on-road sensing information is calibrated and aggregated.

\subsection{Model-Agnostic Fusion Pipeline}%
Agents with distinct perception models usually generate systematically different confidence. The mismatch in confidence distributions can affect the fusion performance. For instance, an inferior model may be over-confident and dominate the aggregation process, decreasing the accuracy of the final results.

To address the issue, we train a calibrator offline for each model, aligning its confidence score with its empirical accuracy on a calibration dataset. First, each model runs its well-trained detector on a public dataset to produce a model-specific dataset with labels and confidence scores. The public dataset, like nuScenes~\cite{nuscenes2019} or Waymo open dataset~\cite{sun2020scalability}, should be independent of the manufacturer and sensor setup, serving only to test the model's performance. The calibration dataset is then used to train the calibrator (see \cref{sec:confidence_calibration} for more details). After training, the calibrator is saved locally for each agent.

When the vehicle is driving on-road and making predictions from the sensor measurements, the calibrator will align the predicted confidence score towards the same standard, thus alleviating the aforementioned mismatch. Then the bounding box coordinates and calibrated confidence scores are packed together and transmitted to neighboring agents. The receiving agent (i.e., ego vehicle) will fuse the shared information via the Promote-Suppress Aggregation algorithm~(see \cref{sec:PSA} for details) to output the final results. Since each agent learns its calibrator independently in the offline stage and only shares the detection outputs during the online phase, the detector architecture and parameters are invisible to other agents, protecting the intellectual property.

\label{sec:system_design}

\subsection{Classification Confidence Calibration}%
\label{sec:confidence_calibration}
To eliminate the bias brought by the system heterogeneity, the models need to be \emph{well-calibrated}. If the confidence scores can imply the likelihood of correct prediction, for example, $80\%$ confidence leads to $80\%$ accurate predictions, this model is \emph{well-calibrated}. Formally, let $\tilde{s}$ be the confidence score produced by the model and $y\in\{0,1\}$ be the label indicating vehicle or background\footnote{We discuss binary classification here for simplicity but the proposed framework can be generalized to the multi-class case.}. A model is \textit{well-calibrated} if its confidence score $\tilde{s}$ matches the expectation of correctly predicting the label:
\begin{equation}
    \mathbb{E}[y=1 \mid \tilde{s}]=\tilde{s}.
    \label{eq:well_calibrated}
\end{equation}

\noindent\textbf{Scaling-Based Confidence Calibration.}
The goal of scaling-based confidence calibration is to learn a parametric scaling function (i.e., calibrator) $\mathtt{c}_{\bm{\theta}}(\tilde{s}):[0,1]\mapsto[0,1]$ on a calibration dataset to transform the uncalibrated confidence scores $\tilde{s}$ into well-calibrated ones $s$. Given a calibration set $\mathcal{D}\triangleq\{(\tilde{s}_{n},y_{n})\}_{n=1}^{N}$ containing the model-dependent confidence scores $\tilde{s}$ and ground-truth labels $y$, we optimize the parameters $\bm{\theta}$ of the calibrator $\mathtt{c}_{\bm{\theta}}(\tilde{s})$ by gradient descent on the binary cross entropy loss
\begin{equation}
  \ell_{CE} = -y_{n}\log(s_{n})-(1-y_{n})\log\left(1-s_{n}\right),
  \label{eq:bce_loss}
\end{equation}
where $s_{n}=\mathtt{c}_{\bm{\theta}}(\tilde{s}_{n})$. Training a parametric function by optimizing \cref{eq:bce_loss} is similar to standard binary classification, however, extra constraints are required  on the scaling function for confidence calibration . Designing a suitable calibrator for our application requires satisfying three conditions: (a) The scaling function needs to be \textit{monotonically non-decreasing} as a higher confidence score is supposed to indicate a higher expected accuracy; (b) The scaling function should be relatively smooth to avoid over-fitting to the calibration set; (c) The scaling function is supposed to be \textit{doubly bounded}, meaning that it maps a confidence interval $[0,1]$ to the same $[0,1]$ range. In the following sub-sections, we will explain why the commonly used calibration methods do not meet all these conditions, which motivates the development of our proposed calibrator.

\begin{figure}[t]%
  \centering%
  \subfloat[Logistic\label{fig:logistic}]{%
    \resizebox{0.4\linewidth}{!}{
      \begin{tikzpicture}%
        \node at(0.0,0.0){\includegraphics[width=\linewidth,trim={3 3 3 3},clip]{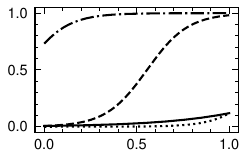}};%
      \end{tikzpicture}%
    }%
  }%
  \subfloat[Kumaraswamy\label{fig:kumaraswamy}]{%
    \resizebox{0.4\linewidth}{!}{
      \begin{tikzpicture}%
        \node at(0.0, 0.0){\includegraphics[width=\linewidth,trim={3 3 3 3},clip]{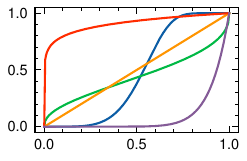}};%
      \end{tikzpicture}%
    }%
  }%
  \caption{\textbf{Scaling functions with various parameters} that follow (a) the logistic form and (b) the Kumaraswamy CDF. Note that, in (b), the ``inverse-sigmoid'' shape (green curve, $a=0.4,b=0.4$) and the identity map (orange curve, $a=1,b=1$) are not in the logistic family.}%
  \label{fig:logistic_kumaraswamy}%
\end{figure}

\textbf{Platt Scaling and Temperature Scaling.}
The most popular scaling methods are arguably Platt Scaling~\cite{platt1999probabilistic} and Temperature Scaling~\cite{guo2017calibration}.
Platt Scaling uses the logistic family as the calibrator:
\begin{equation}
  \mathtt{c}_{\text{Platt}}(\tilde{s};a,b)=\frac{1}{1+\exp\left(-(a\times{\tilde{s}}+b)\right)},
  \label{eq:logistic}
\end{equation}
where $a,b$ are parameters with $a\geq{0}$ to ensure that the calibration map is \textit{monotonically non-decreasing}. Temperature Scaling is a special case of \cref{eq:logistic} where $b$ is fixed to $0$. \cref{fig:logistic_kumaraswamy} shows several scaling functions from this family. Platt Scaling can fail if its parametric assumptions are not met~\citep{kull2017beta}. For example, we cannot learn an ``inverse-sigmoid'' (see the green curve in \cref{fig:kumaraswamy}) scaling function within this family. Furthermore, the identity function is also not a member of the logistic family. In addition to the aforementioned limitations, the logistic family is also not a function family that can naturally map $[0,1]$ to $[0,1]$ as its input domain is $\mathbb{R}$, therfore, these popular choices are not our ideal candidates.

\textbf{Doubly Bounded Scaling (DBS).}
We propose to use the Kumaraswamy Cumulative Density Function (CDF)~\cite{kumaraswamy1980generalized} that meets all the three constraints while being sufficiently flexible. To the best of our knowledge, this is the first time that this function family has been adopted in confidence calibration.
Specifically, we learn a scaling function with the following form
\begin{equation}
  \mathtt{c}(\tilde{s};a,b)=1-\left(1-\tilde{s}^{a}\right)^{b},
  \label{eq:dbs}
\end{equation}
where $a>0$ and $b>0$ are the parameters. Scaling functions that follow \cref{eq:dbs} are \textit{monotonically non-decreasing}, \textit{smooth}, and \textit{doubly bounded}, hence the name.
we can see that DBS is more flexible than the logistic form by comparing \cref{fig:logistic} and \cref{fig:kumaraswamy}.
For each detector, we optimize the $a$ and $b$ on a calibration dataset by minimizing \cref{eq:bce_loss}. 

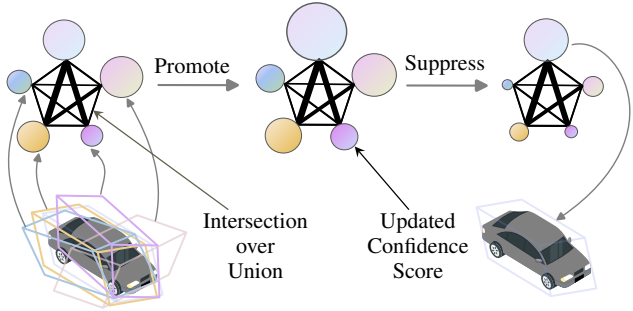
\begin{figure}[t]%
  \centering%
  \subfloat{%
    \resizebox{\linewidth}{!}{
      \scriptsize
      \begin{tikzpicture}%
        \node at(0.0,0.0){\includegraphics[width=0.8\linewidth]{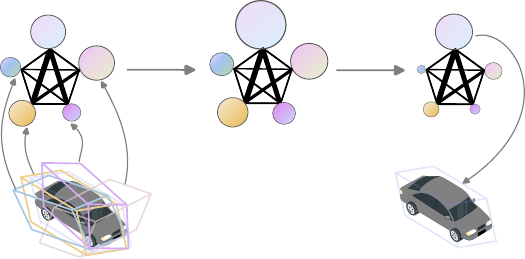}};%
        \node at (-1.4,1) {Promote};
        \node at (1.4,1) {Suppress};
        \draw [-stealth,gray] (-1,-0.6) -- (-2.5,0.5);
        \node [align=center] at (-0.7,-1) {Intersection\\over\\Union};
        \draw [-stealth,black] (1,-0.6) -- (0.4,0.1);
        \node [align=center] at (1.1,-1) {Updated\\Confidence\\Score};
      \end{tikzpicture}%
    }%
  }\vspace{-12pt}%
  \caption{\textbf{Illustration of Promote-Suppress Aggregation}. The size of a node indicates the confidence score of the bounding box and the edge width represents the Intersection-over-Union of two boxes.}\label{fig:psa}%
\end{figure}
\subsection{Promote-Suppress Aggregation (PSA)}\label{sec:PSA}
Detection models typically output a bunch of overlapped bounding box candidates for the same detected object, thus we need a post-processing step to select from these candidates. In most of the detection algorithms, the optimization objective function is a summation of a bounding box regression loss and a classification loss. The detector can express its ``confidence'' by assigning high classification scores to the promising bounding boxes or allocating more bounding boxes to the region that it finds relevant features. To select the high-score bounding boxes with many confident neighbors, we propose Promote-Suppress Aggregation~(PSA), which takes into account both the regression and classification confidences.

\cref{fig:psa} illustrates the idea of PSA. We first construct a spatial graph of bounding box candidates based on Intersection-over-Union (IoU)  values and the confidence scores. In the promotion step, the IoU weighted confidence scores are propagated to the neighboring nodes. We design the propagation rule to meet the following desiderata:
\begin{itemize}
  \item A candidate should be promoted if many other candidate boxes have \textit{large intersections} with it;
  \item A candidate with many \textit{high-score neighbors} should be promoted;
  \item If possible, the update rules should be \textit{parallelizable} and \textit{permutation-invariant}. Namely, the propagation order does not change the result.
\end{itemize}
In the suppression step, the candidate with the highest updated score will softly suppress the scores of other candidates. Finally, we select one or more bounding boxes that rank in the first (few) places. The idea of soft suppression and selecting more than one candidate is akin to soft-NMS~\citep{bodla2017soft}, which is beneficial when the bounding box of a small object is within the box of a large object. Below we formally describe the PSA algorithm.

\begin{definition}[Bounding Box Graph]
Let $\mathcal{G}$ be a weighted graph with a set of edges $\mathcal{E}$ and a set of nodes/vertices $\mathcal{V}$, where each vertex $v\in\mathcal{V}$ represents a bounding box candidate $b$ with an associated confidence score $s$ after calibration. The edge weigh $w_{ij}$ between vertex $v_{i}$ and $v_{j}$ is defined as the Intersection-over-Union value $\mathtt{IoU}(b_{i},b_{j})\triangleq\nicefrac{\cap(b_{i},b_{j})}{\cup(b_{i},b_{j})}$. An edge connects vertex $v_{i}$ and $v_{j}$ if the edge weight is non-zero.
\end{definition}

\begin{definition}[Connected Components]
The graph consists of a number of \textit{connected components} in which every pair of nodes is connected via a sequence of edges.
\end{definition}

\begin{problem}[Bounding Box Aggregation]
Given the Intersection-over-Union matrix $\mathbf{U}\in[0,1]^{N\times{N}}$ among $N$ bounding box candidates $\mathbf{B}=[\mathbf{b}_{1},\dots,\mathbf{b}_{N}]^{\T}$ and their confidence scores $\mathbf{s}=[s_{1},\dots,s_{N}]^{\T}$, our goal is to compute an index set $\mathcal{I}$ to select/filter candidates that best match the ground-truth bounding boxes.
\end{problem}

{\algrenewcommand\textproc{}
  \begin{algorithm}[t]\small
    \setstretch{1.3}
    \caption{\textbf{Promote-Suppress Aggregation}}\label{alg:psa}
    \hspace*{\algorithmicindent} \textbf{Arguments}: bounding boxes $\mathbf{B}=[\mathbf{b}_{1},\dots,\mathbf{b}_{N}]^{\T}$,\\
    \hspace*{\algorithmicindent}\hspace{48pt} confidence score vector $\mathbf{s}=[s_{1},\dots,s_{N}]^{\T}$,\\
    \hspace*{\algorithmicindent}\hspace{48pt} soft selection parameters $\varepsilon$, and threshold $\phi$
    \begin{algorithmic}[1] 
      \State Initialize selected box indices to an empty set $\mathcal{I}=\emptyset$
      \State Compute IoU matrix $\mathbf{U}\in[0,1]^{N\times{N}}$ using $\mathbf{B}$
      \State Find vertex indices of connected components $\mathcal{C}\triangleq\{\mathbf{c}_{m}\}_{m=1}^{M}$
      \For {each $\mathbf{c}_{m}\in\mathcal{C}$}
        \State Extract IoU sub-matrix $\mathbf{U}_{m}\in[0,1]^{N_{m}\times{N_{m}}}$ via $\mathbf{c}_{m}$
        \State Extract score sub-vector $\mathbf{s}_{m}\in[0,1]^{N_{m}}$ via $\mathbf{c}_{m}$
        \State $\mathbf{\hat{s}}_{m}=\mathbf{U}_{m}\mathbf{s}_{m}$\Comment{Promote}
        \State $\mathbf{\bar{s}}_{m}=\mathtt{softmax}(\nicefrac{\mathbf{\hat{s}}_{m}}{\varepsilon})$\Comment{Suppress}
        \State $\mathcal{I}=\mathcal{I}\cup\{c_{m}^{(n)} \mid \bar{s}_{m}^{(n)}>\phi, n=1,\dots,N_m\}$\Comment{Select}
      \EndFor
      \State \textbf{return} selected candidate indices $\mathcal{I}$
      \end{algorithmic} 
    \end{algorithm}
  }

Algorithm\,\ref{alg:psa} shows how PSA computes the index set. Given the IoU adjacency matrix, we can find out the indices of each component and put them into a component set $\mathcal{C}=\{\mathbf{c}_{m}\}_{m=1}^{M}$, where $M$ is the number of components and $\mathbf{c}_{m}$ contains the indices of $N_{m}$ vertices (line 3).
For each component, we extract the IoU matrix $\mathbf{U}_{m}\in[0,1]^{N_{m}\times{N_{m}}}$ and confidence score vector $\mathbf{s}_{m}\in[0,1]^{N_{m}}$ corresponding to this component (line 5-6).
Then, we perform the promotion step $\mathbf{\hat{s}}_{m}=\mathbf{U}_{m}\mathbf{s}_{m}$ where each vertex updates its score to be the IoU-weighted sum of scores from other vertices in the component (line 7).  In the suppression step, we normalize the updated scores back to $[0,1]$ and distill the winning candidate via $\mathbf{\bar{s}}_{m}=\mathtt{softmax}(\nicefrac{\mathbf{\hat{s}}_{m}}{\varepsilon})$ (line 8).
In the end, indices with updated scores larger than a threshold are added to the set $\mathcal{I}$ (line 9).
We can select multiple candidates if $\varepsilon\in(0,1]$ is large and $\phi$ is small.
In our application, however, one component typically contains only one object/vehicle, so we use a small $\varepsilon$ and $\phi=0.5$.
Overall, PSA is highly parallelizable as each component operates independently and each step only requires simple linear search or small matrix-vector multiplication.
\section{Experiments}%
\label{sec:experiments}
\subsection{Dataset}
We evaluate the proposed framework on a large-scale open-source multi-agent perception dataset OPV2V~\cite{xu2022opencood}, which is simulated using the high-fidelity simulator CARLA~\cite{dosovitskiy2017carla} and a cooperative driving automation simulation framework OpenCDA~\cite{xu2021opencda}. It includes $73$ scenarios with an average of $25$ seconds duration. In each scene, various numbers ($2$ to $7$) of Autonomous Vehicles (AVs) provide LiDAR point clouds from their viewpoints. The train/validation/test splits are $6764/1981/2169$ frames, respectively. For details of the dataset, please refer to~\cite{xu2022opencood}.

\begin{table}[t]
  \centering
  \scriptsize
  \caption{Object detection performance. Average Precision~(AP) at IoU=0.7 on \textit{Homo}, \textit{Hetero1}, and \textit{Hetero2} setting.}
  \begin{tabular}{lccc}
    \toprule
    \textbf{Methods} &
    \multicolumn{1}{c}{\begin{tabular}[c]{@{}c@{}}\textbf{Homo}\\    $\uparrow$AP@0.7\end{tabular}} &
    \multicolumn{1}{c}{\begin{tabular}[c]{@{}c@{}}\textbf{Hetero1}\\ $\uparrow$AP@0.7\end{tabular}} &
    \multicolumn{1}{c}{\begin{tabular}[c]{@{}c@{}}\textbf{Hetero2}\\ $\uparrow$AP@0.7\end{tabular}} \\
    \midrule
    No fusion                    & 0.602          & 0.602          & 0.602          \\
    Intermediate w/o calibration & \textbf{0.815} & 0.677          & 0.571          \\
    Late fusion w/o calibration  & 0.781          & 0.691          & 0.723          \\
    Our method                   & 0.813          & \textbf{0.750} & \textbf{0.784} \\
    \bottomrule
  \end{tabular}
  \label{table:main_results}
\end{table}



\subsection{Experiment Setup}
\noindent\textbf{Evaluation metric.} Following~\cite{xu2021opencda}, we evaluate the detection accuracy in the range of $x\in[-140, 140]\text{m}$ and $y\in[-40, 40]\text{m}$, centered at the ego-vehicle coordinate frame. The detection performance is measured with Average Precision~(AP) at $IoU=0.7$. 

\noindent\textbf{Evaluation setting.} We evaluate our method under three different settings: 1) \textit{Homo Setting}, where the detectors of agents are homogeneous with the same architecture and trained parameters. This setting has no confidence distribution gap and is used to demonstrate the performance drop when taking heterogeneity into account;  2) \textit{Hetero Setting~1}, where the agents have the same model architecture but different parameters; 3) \textit{Hetero Setting~2}, where the detector architectures are disparate. For \textit{Homo Setting}, we select pre-trained Pointpillar~\cite{pointpillar} as the backbone for all the AVs. For \textit{Hetero Setting~1}, the ego vehicle employs the same pre-trained Pointpillar model as in \textit{Homo Setting}, whereas other AVs pick the parameters of Pointpillar from a different epoch during training. Likewise, in the \textit{Hetero Setting~2},  the ego vehicle utilizes Pointpillar while other AVs use SECOND~\cite{Yan2018SECONDSE} for detection. As intermediate fusion requires equal feature map resolution,  we apply simple bi-linear interpolation under this setting. The ego vehicle uses the identical model with the same parameters across all settings for the \textit{No Fusion} and \textit{Late Fusion}. To compare with existing calibrators, we use the same calibration method for all agents, but the parameters are agent-specific. The proposed framework should also work even when the calibration methods across agents are heterogeneous, as long as the prediction bias is effectively reduced.

\noindent\textbf{Compared methods.}
We regard \textit{No Fusion} as the baseline, which only takes the ego vehicle's LiDAR data as input and omits any collaboration. Ideally, the multi-agent system should at least outperform this baseline. To validate the necessity of the calibration, we compare our method with naive late fusion and intermediate fusion that ignore calibrations. The naive late fusion gathers all detected bounding box positions and confidence scores together and simply applies NMS to produce the final results. The intermediate fusion method is the same as the one in~\cite{xu2022opencood}. We exclude the early fusion in the comparison as it requires large bandwidth, which leads to high communication delay thus is impractical to be deployed in the real world.
Moreover, we also compare the proposed Doubly Bounded Scaling (DBS) with two other commonly used scaling-based calibrators: Temperature Scaling (TS)~\cite{guo2017calibration} and Platt Scaling (PS)~\cite{platt1999probabilistic}.

\begin{table}[t]
  \centering
  \scriptsize
  \caption{Component ablation study.}
  \begin{tabular}{cccc}
    \toprule
   \multicolumn{2}{c}{\textbf{Components}}& \textbf{Hetero1} & \textbf{Hetero2}\\
    DBS & PSA & $\uparrow$AP@0.7 & $\uparrow$AP@0.7\\
    \midrule
     &  & 0.691 & 0.723\\
    \cmark &  & 0.734 & 0.776\\
    \cmark & \cmark & \textbf{0.750} & \textbf{0.784}\\
    \bottomrule
  \end{tabular}
  \label{table:ablation}
\end{table}

\subsection{Quantitative Evaluation}\label{sec:results}%
\noindent \textbf{Main performance analysis.}  \cref{table:main_results} describes the performance comparisons of different methods under \textit{Homo}, \textit{Hetero1}, and \textit{Hetero2 Setting}. In the unrealistic \textit{Homo} setting, all methods exceed the baseline remarkably while intermediate fusion and our method have very close performance~(0.2\% difference). However, when we consider the realistic model discrepancy factor, our method outperforms the classic late fusion and intermediate fusion significantly by 5.9\%,  7.3\% under \textit{Hetero1 Setting}, and by 6.1\%, 21.3\% under \textit{Hetero2 Setting}, respectively. The classic late fusion and intermediate fusion suffer from the model discrepancy, leading to clear accuracy decreases. In the \textit{Hetero2 Setting}, the intermediate fusion even becomes lower than the baseline.
On the contrary, our method only drops around 6\% and 3\% under the two realistic settings, indicating the effectiveness of the proposed calibration for the heterogeneity of the multi-agent perception system. Note that although the design essence of our framework aims to handle the heterogeneous situations, we also obtain performance boost under the ~\textit{Homo Setting} compared with the standard late fusion that shares detection proposals. We attribute this gain to PSA and the filtering operation of low-confidence proposals after confidence calibration that removes some potential false positives.

\noindent \textbf{Major component analysis.}
Here we investigate the contribution from each component by incrementally adding DBS and PSA.
\cref{table:ablation} reveals that both modules are beneficial for the performance boost, while the calibration exhibits more contributions -- increasing the AP by 4.3\% and 5.3\% .

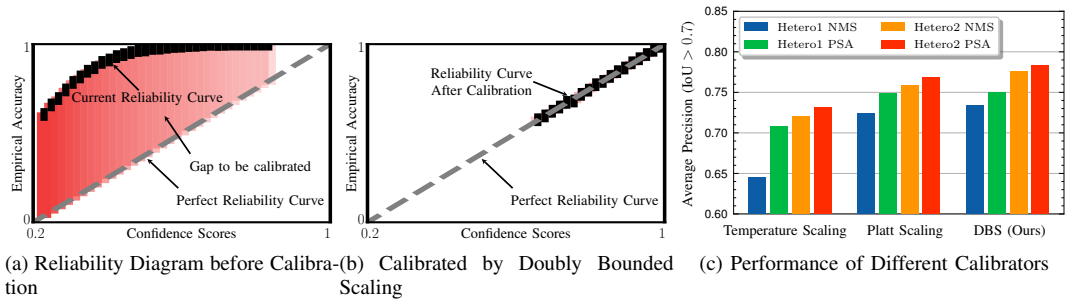
\begin{figure*}[t]%
  \centering%
  \subfloat[Reliability Diagram before Calibration\label{fig:before_dbs}]{%
    \resizebox{0.25\linewidth}{!}{
      \begin{tikzpicture}\Huge%
        \node at(0.0,0.0){\includegraphics[width=\linewidth,height=0.6\linewidth,trim={3 3 3 3},clip]{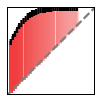}};%
        \draw [-stealth, line width=1mm,](-0.2,-3.2) -- (-2,-1.5);
        \node at(4,-4) {Perfect Reliability Curve};%
        \draw [-stealth, line width=1mm,](0.5,-1.2) -- (-1,1);
        \node at(4,-2) {Gap to be calibrated};%
        \draw [-stealth, line width=1mm,](-2.5,2.5) -- (-4,4);
        \node at(-2,2) {Current Reliability Curve};%
        \node at(-9, -5) {$0$};%
        \node at(-9,5) {$1$};%
        \node[rotate=90] at (-9.5,0.0) {Empirical Accuracy};%
        \node at(-8.5,-5.8) {$0.2$};%
        \node at(8.5,-5.8) {$1$};%
        \node at(0,-5.8) {Confidence Scores};%
      \end{tikzpicture}%
    }%
  }%
  \subfloat[Calibrated by Doubly Bounded Scaling\label{fig:after_dbs}]{%
    \resizebox{0.25\linewidth}{!}{
      \begin{tikzpicture}\Huge%
        \node at(0.0,0.0){\includegraphics[height=0.6\linewidth,width=\linewidth,trim={3 3 3 3},clip]{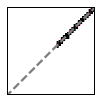}};%
        \draw [-stealth, line width=1mm,](1,3) -- (3,2.2);
        \node [align=center] at(-2,3) {Reliability Curve\\After Calibration};%
        \draw [-stealth, line width=1mm,](-0.2,-3.2) -- (-2,-1.5);
        \node at(4,-4) {Perfect Reliability Curve};%
        \node at(-9, -5) {$0$};%
        \node at(-9,5) {$1$};%
        \node[rotate=90] at (-9.5,0.0) {Empirical Accuracy};%
        \node at(-8.5,-5.8) {$0.2$};%
        \node at(8.5,-5.8) {$1$};%
        \node at(0,-5.8) {Confidence Scores};%
      \end{tikzpicture}%
    }%
  }%
  \subfloat[Performance of Different Calibrators\label{fig:calibrators}]{%
    \resizebox{0.30\linewidth}{!}{
      \begin{tikzpicture}\LARGE%
        \node at(0.0,0.0){\includegraphics[width=0.8\linewidth]{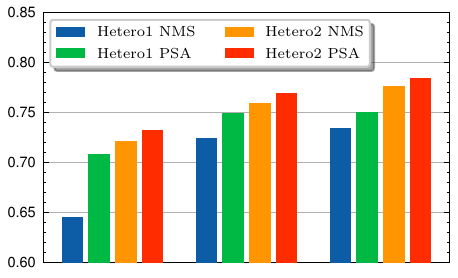}};%
        \node at (-3.8,-4.5) {Temperature Scaling};%
        \node at (0.8,-4.5) {Platt Scaling};%
        \node at (4.8,-4.5) {DBS (Ours)};%
        \node[rotate=90] at (-7.5,0.0) {Average Precision ($\text{IoU}>0.7$)};%
      \end{tikzpicture}%
    }%
  }%
  \caption{The reliability diagrams in (a) and (b) reveal that Doubly Bounded Scaling method can effectively calibrate the classification confidence scores. In (c), the proposed Doubly Bounded Scaling outperforms Temperature Scaling and Platt Scaling under various experiment setups and aggregation algorithms.}\label{fig:calibration}\vspace{-10pt}%
\end{figure*}

\begin{figure*}[t]%
  \centering%
  \subfloat[Intermediate Fusion in Hetero1]{%
    \resizebox{0.25\linewidth}{!}{
      \begin{tikzpicture}%
        \node at(0.0,0.0){\fbox{\includegraphics[width=\linewidth]{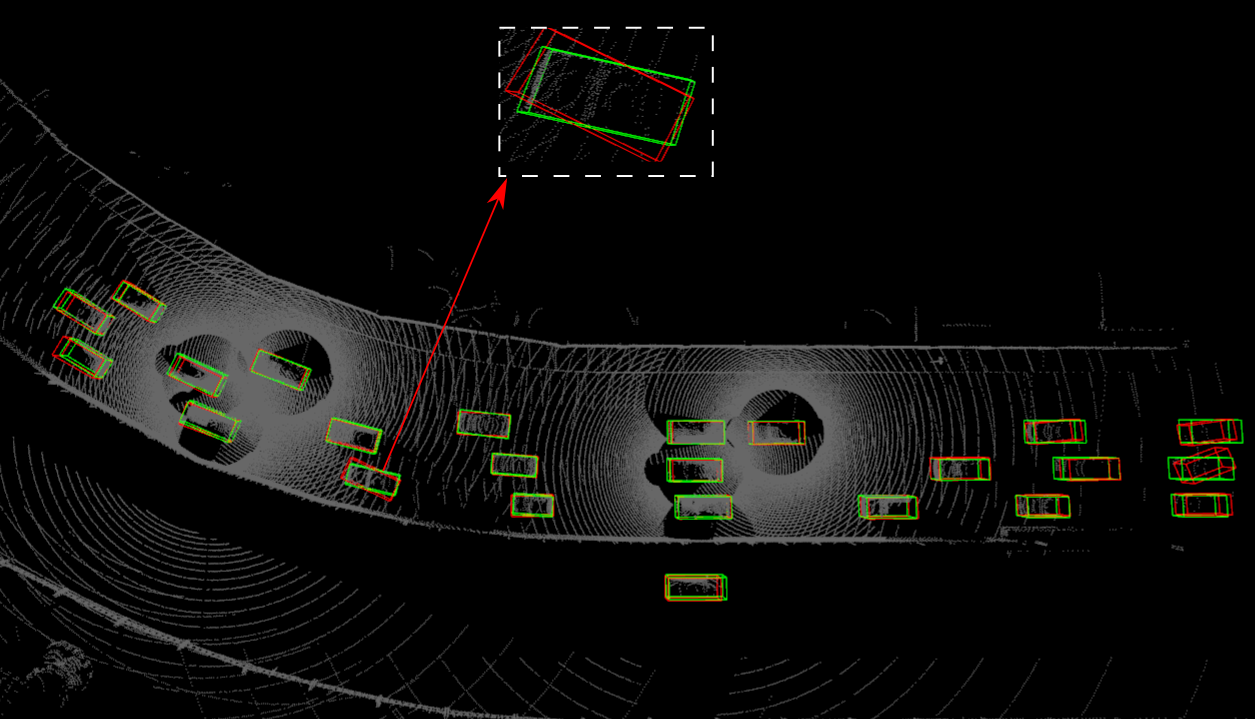}}};%
      \end{tikzpicture}%
    }%
  }%
  \subfloat[Late Fusion in Hetero1]{%
    \resizebox{0.25\linewidth}{!}{
      \begin{tikzpicture}%
        \node at(0.0,0.0){\fbox{\includegraphics[width=\linewidth]{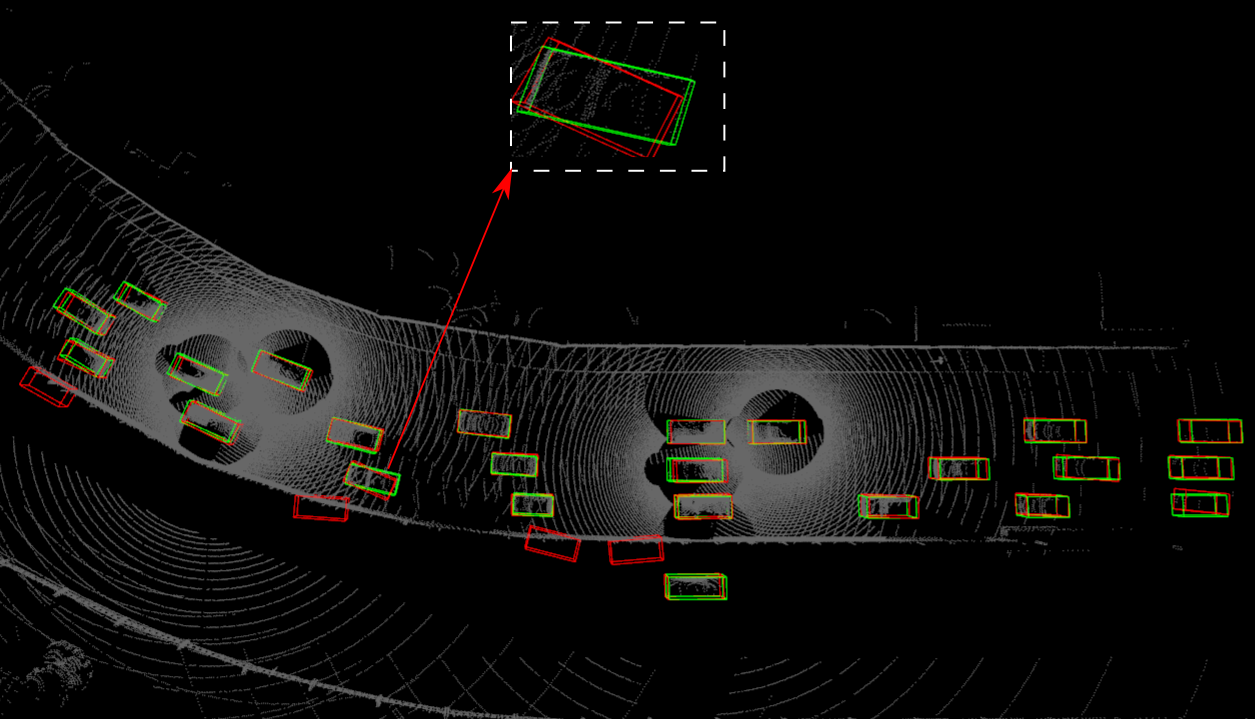}}};%
      \end{tikzpicture}%
    }%
  }%
  \subfloat[Ours in Hetero1]{%
    \resizebox{0.25\linewidth}{!}{
      \begin{tikzpicture}%
        \node at(0.0,0.0){\fbox{\includegraphics[width=\linewidth]{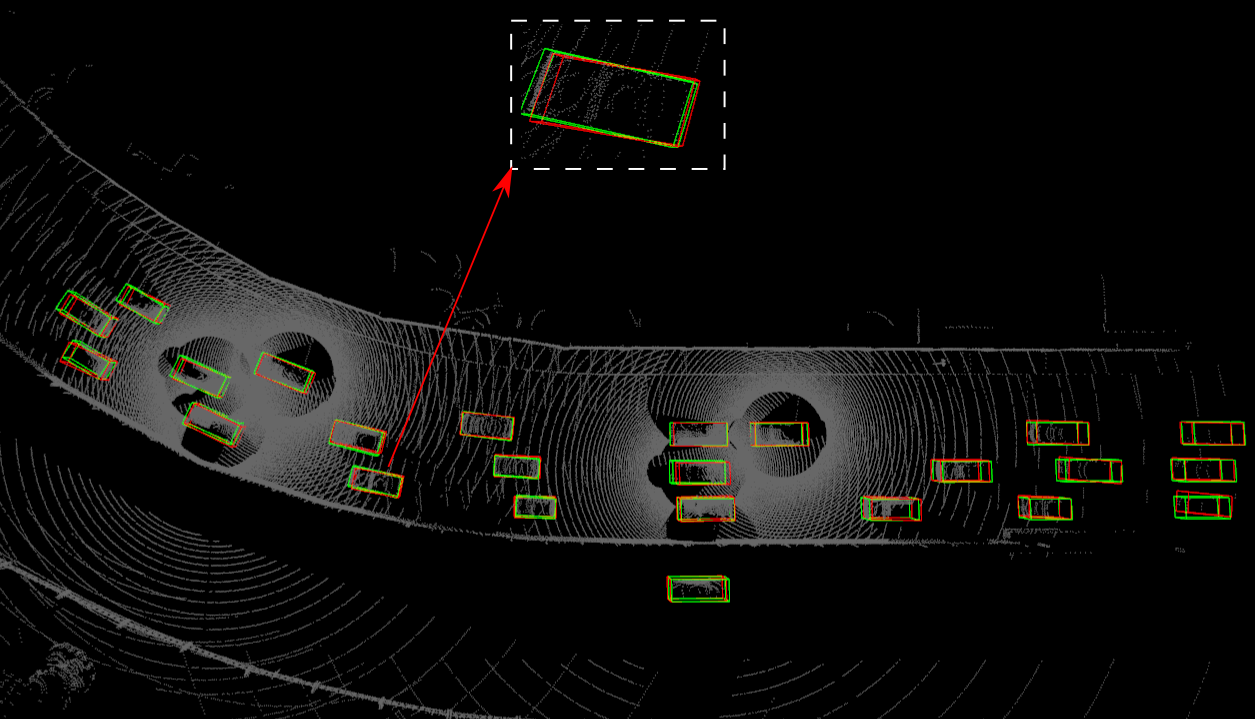}}};%
      \end{tikzpicture}%
    }%
  }\vspace{-8pt}\\%
  \subfloat[Intermediate Fusion in Hetero2]{%
    \resizebox{0.25\linewidth}{!}{
      \begin{tikzpicture}%
        \node at(0.0,0.0){\fbox{\includegraphics[width=\linewidth]{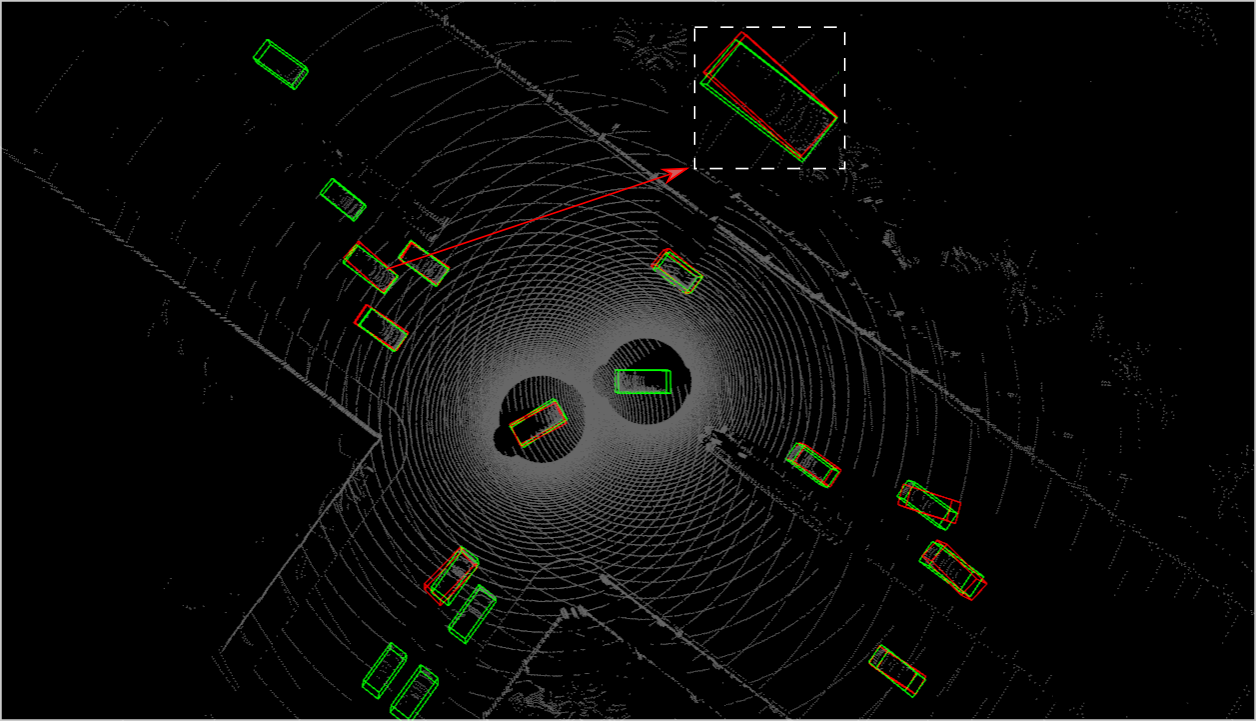}}};%
      \end{tikzpicture}%
    }%
  }%
  \subfloat[Late Fusion in Hetero2]{%
    \resizebox{0.25\linewidth}{!}{
      \begin{tikzpicture}%
        \node at(0.0,0.0){\fbox{\includegraphics[width=\linewidth]{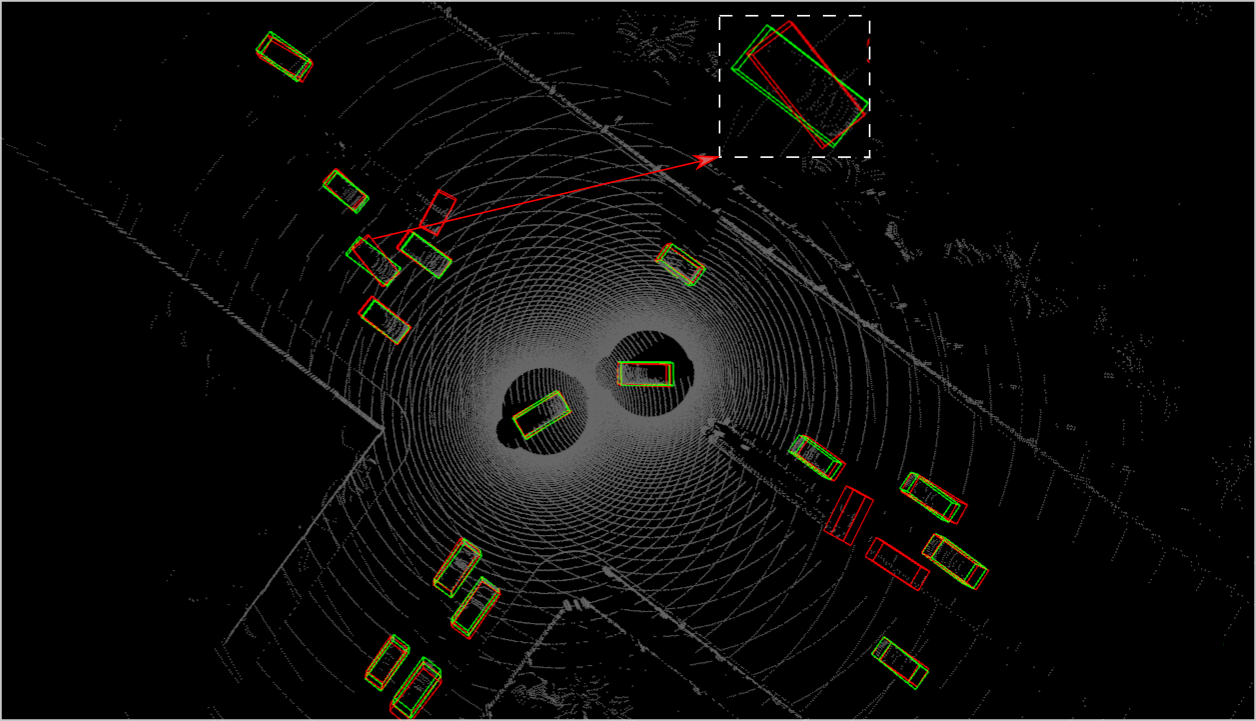}}};%
      \end{tikzpicture}%
    }%
  }%
  \subfloat[Ours in Hetero2]{%
    \resizebox{0.25\linewidth}{!}{
      \begin{tikzpicture}%
        \node at(0.0,0.0){\fbox{\includegraphics[width=\linewidth]{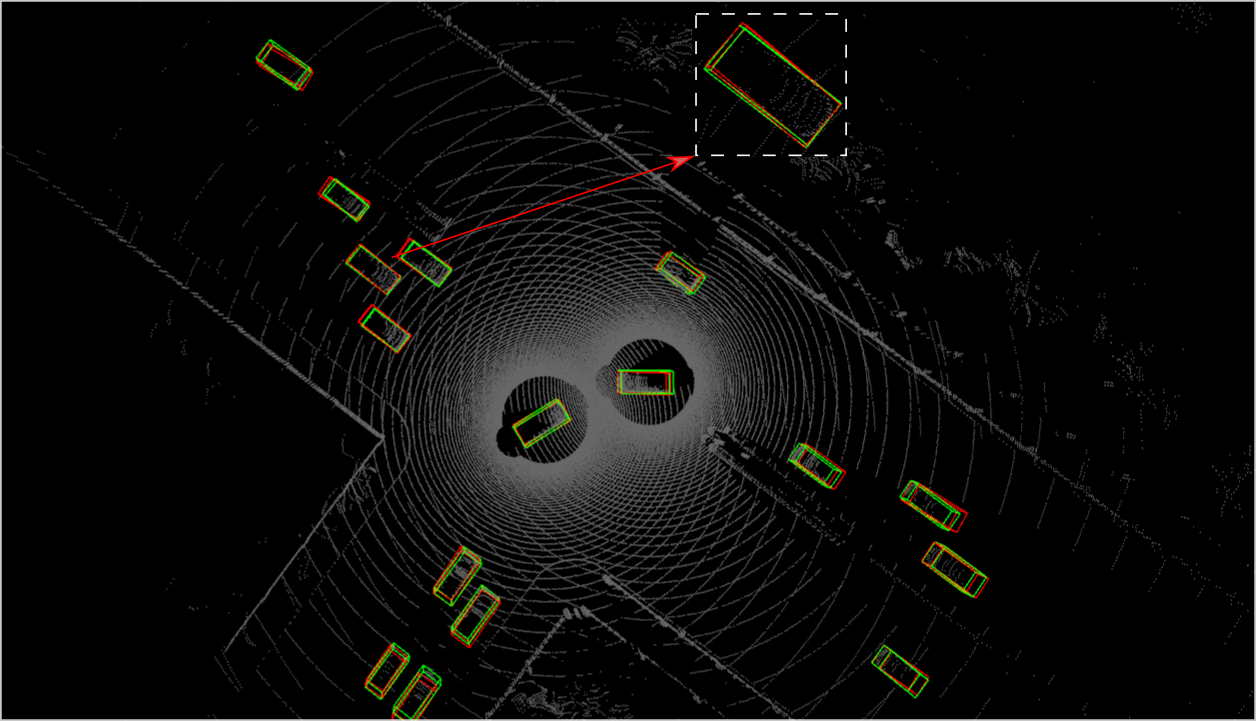}}};%
      \end{tikzpicture}%
    }%
  }%
  \caption{\textbf{Qualitative comparison in a busy freeway and a congested intersection.} \textcolor{green}{Green} and \textcolor{red}{red} 3D bounding boxes represent the groun truth and prediction respectively. Our method yields more accurate detection results.}%
  \label{fig:visual}\vspace{-10pt}%
\end{figure*}

\noindent\textbf{Confidence calibration evaluation.}
\cref{fig:before_dbs} show the reliability diagram of Pointpillar used by the ego vehicle, in which a perfect calibration will produce a diagonal reliability curve, indicating the real accuracy matches the predictive confidence score. Reliability curves under or above the diagonal line represent over-confident or under-confident models, respectively. Pointpillar has much higher empirical accuracy than its reported confidence score. When using NMS to fuse the predictions of Pointpillar with that of another inaccurate but over-confident detector, the under-estimated confidence will result in the removal of Pointpillar's good predictions. After being calibrated by DBS, in \cref{fig:after_dbs}, the reliability curve of Pointpillar lies on the diagonal line.

\noindent\textbf{Comparison with other calibration methods.}
\cref{fig:calibrators} describes the comparison between our DBS calibration and other calibration methods, including TS and PS. Our DBS achieves better performance than others under both heterogeneous settings. Moreover, PSA can also improve the accuracy of different calibrators and experimental settings, showing the generalized capability to refine the prediction results.

\subsection{Qualitative Results}
\cref{fig:visual} shows the detection results of intermediate fusion, classic late fusion, and our method under \textit{Hetero1} and \textit{Hetero2 Setting}. Our method can identify more objects while keeping very few false positives. The zoom-in examples show that our method can regress the bounding box positions more accurately, indicating the robustness against the model discrepancy in multi-agent perception systems. 

\section{Conclusions}%
\label{sec:conclusions}
In the context of cooperative perception, agents from different stakeholders have heterogeneous models. For the sake of confidentially, information related to the models and parameters should not be revealed to other agents.
In this work, we present a model-agnostic collaboration framework that addresses two critical challenges of the vanilla late fusion strategy. First, we propose a confidence calibrator to align the classification confidence distributions of different agents. Second, we present a bounding box aggregation algorithm that takes into account both the calibrated classification confidence and the spatial congruence information given by bounding box regression. Experiments on a large-scale cooperative perception dataset shed light on the necessity of model calibration across heterogeneous agents.
The results show that combining the two proposed techniques can improve the state-of-the-art for cooperative 3D object detection when different agents use distinct perception models.

\section{Acknowledgment}%
This work is part of the OpenCDA Ecosystem~\cite{10045043} and supported in part by the Federal Highway Administration Exploratory Advanced
Research (EAR) Program,

\bibliographystyle{unsrtnat}
\balance
\bibliography{references}
\end{document}